# Exploring the Decision Forest: An Empirical Investigation of Occam's Razor in Decision Tree Induction

**Patrick M. Murphy**                                    PMURPHY@ICS.UCI.EDU
**Michael J. Pazzani**                                   PAZZANI@ICS.UCI.EDU
*Department of Information & Computer Science*
*University of California, Irvine, CA 92717*

## Abstract

We report on a series of experiments in which all decision trees consistent with the training data are constructed. These experiments were run to gain an understanding of the properties of the set of consistent decision trees and the factors that affect the accuracy of individual trees. In particular, we investigated the relationship between the size of a decision tree consistent with some training data and the accuracy of the tree on test data. The experiments were performed on a massively parallel Maspar computer. The results of the experiments on several artificial and two real world problems indicate that, for many of the problems investigated, smaller consistent decision trees are on average less accurate than the average accuracy of slightly larger trees.

## 1. Introduction

The top-down induction of decision trees is an approach to machine learning that has been used on a variety of real world tasks. Decision trees are well-suited for such tasks since they scale fairly well with the number of training examples and the number of features, and can represent complex concepts in a representation that is fairly easy for people to understand.

Decision tree induction algorithms (Breiman, Friedman, Olshen, & Stone, 1984; Quinlan, 1986; Fayyad & Irani, 1992) typically operate by choosing a feature that partitions the training data according to some evaluation function (e.g., the purity of the resulting partitions). Partitions are then further partitioned recursively until some stopping criterion is reached (e.g., the partitions contain training examples of a single class). Nearly all decision tree induction algorithms create a single decision tree based upon local information of how well a feature partitions the training data. However, this decision tree is only one of a set of decision trees consistent with the training data. In this paper, we experimentally examine the properties of the set of consistent decision trees. We will call the set of decision trees that are consistent with the training data a *decision forest*.

Our experiments were run on several artificial concepts for which we know the correct answer and two naturally occurring databases from real world tasks available from the UCI Machine Learning Repository (Murphy & Aha, 1994) in which the correct answer is not known. The goal of the experiments were to gain insight into how factors such as the size of a consistent decision tree are related to the error rate on classifying unseen test instances. Decision tree learners, as well as most other learners, attempt to produce the smallest consistent hypothesis.[1] Occam's razor is often used to justify this bias. Here, we experimentally evaluate this bias towards simplicity by investigating the relationship between the size of a consistent decision tree and its accuracy. If the average error of decision trees with $N$ test nodes is less than the average error of decision trees of size $N + i$ (for $i > 0$), an appropriate bias for a learner attempting to minimize average error would be to return the smallest decision tree it can find within its resource constraints.

In this paper, we restrict our attention to decision trees that are consistent with the training data and ignore issues such as pruning which trade off consistency with the training data and the simplicity of the hypothesis. For the purposes of this paper, a consistent decision tree is one that correctly classifies every training example.[2] We also place two additional constraints on decision trees. First, no discriminator can pass all instances down a single branch. This insures that the test made by the decision tree partitions the training data. Second, if all of the training instances at a node are of the same class, no additional discriminations are made. In this case, a leaf is formed with class label specified by the class of the instances at the leaf. These two constraints are added to insure that the decision trees analyzed in the experiments correspond to those that could be formed by top down decision tree induction algorithms. In this paper, we will not investigate problems that have continuous-valued features or missing feature values.

In Section 2 (and the appendix), we will report on some initial exploratory experiments in which the smallest consistent decision trees tend to be less accurate than the average accuracy of those slightly larger. Section 3 provides results of additional experiments that address this issue. Section 4 addresses the implication of our findings to the policy a learner should take in deciding which of the many consistent hypotheses it should prefer. Section 5 relates this work to previous empirical and theoretical research.

## 2. Initial Experiments

We will investigate the relationship between various tree characteristics and error. In particular, we will look at node cardinality (i.e., the number of internal nodes in a tree) and leaf cardinality (i.e., the total number of leaves in a tree).

It should be noted that even when using a powerful massively parallel computer, the choice of problems is severely constrained by the computational complexity of the task. The number of trees of any node cardinality that might be generated is $O(d^c)$ where $d$ is the number of discriminators and $c$ is the node cardinality. Even on a massively parallel computer, this precluded the use of problems with many features or any continuous-valued features.

The first experiment considered learning from training data in which there are 5 boolean features. The concept learned was $XYZ \lor AB$. This concept was chosen because it was of moderate complexity, requiring a decision tree with at least 8 nodes to represent correctly. With 5 boolean features, the smallest concept (e.g., True) would require 0 test nodes and the largest (e.g., parity) would require 31.

---

1. We say "attempt to produce the smallest consistent hypothesis" because most systems use some form of limited look-ahead or greedy search. As a result, the smallest consistent tree is rarely found.

2. The artificial and natural problems we study here have consistent training sets.

We ran 100 trials, creating a training set by randomly choosing without replacement 20 of the 32 possible training examples and using the remaining 12 examples as the test set. For each trial, every consistent decision tree was created, and we computed the average error rate made by trees with the same node cardinality. Figure 1 plots the mean and 95% confidence interval of these average errors as a function of the node cardinality. Figure 1 also plots the number of trials on which at least one decision tree of a given node cardinality is consistent with the training data.

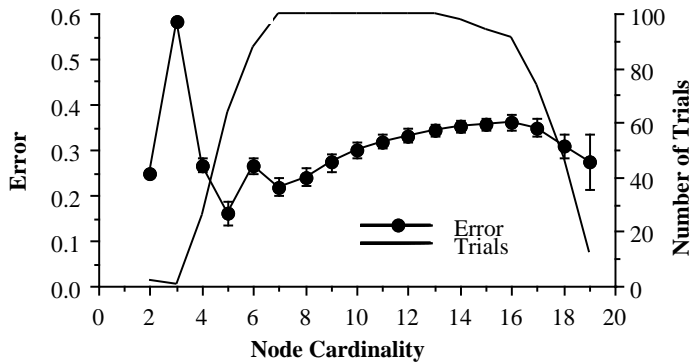

Figure 1. The average error of 100 trials as a function of node cardinality and the number of trials for each node cardinality.

From node cardinality 7 to node cardinality 16, there is a monotonic increase in error with increasing node cardinality. For the range from 2 to 3 nodes, the error is varied; however there is little evidence for these error values because they are based on only 2 and 1 trials, respectively. For the range of node cardinalities between 4 and 7, average error is definitely not a monotonically increasing function of node cardinality. As seen in the curve, 5 node trees are on the average more accurate than 4 node trees, and 7 node trees are on the average more accurate than trees with 6 nodes. This last result is somewhat surprising since one gets the impression from reading the machine learning literature (Muggleton, Srinivasan, & Bain, 1992) that the smaller hypothesis (i.e., the one that provides the most compression of the data (Rissanen, 1978)) is likely to be more accurate. We will explore this issue in further detail in Section 3. Appendix 1 presents data showing that this result is not unique to this particular concept. A final, interesting finding that we will not explore further in this paper is that for very large node cardinalities, error begins to decrease as the node cardinality increases.

Table 1 lists the average number of consistent trees for each node cardinality and the average number of correct trees (i.e., those trees consistent with the training data that make no errors on the unseen test examples). There are no correct trees with fewer than 8 nodes, since at least 8 nodes are required to represent this concept. Clearly, since there are many trees consistent with the training data, a learner needs some policy to decide which tree to return. We will return to this issue in Section 4.

| Nodes | Number of Consistent Trees | Number of Correct Trees |
|-------|----------------------------|-------------------------|
| 2     | 2.0                        | 0.0                     |
| 3     | 4.0                        | 0.0                     |
| 4     | 3.3                        | 0.0                     |
| 5     | 12.3                       | 0.0                     |
| 6     | 27.6                       | 0.0                     |
| 7     | 117.1                      | 0.0                     |
| 8     | 377.0                      | 17.8                    |
| 9     | 879.4                      | 37.8                    |
| 10    | 1799.9                     | 50.2                    |
| 11    | 3097.8                     | 41.6                    |
| 12    | 4383.0                     | 95.4                    |
| 13    | 5068.9                     | 66.6                    |
| 14    | 4828.3                     | 37.7                    |
| 15    | 3631.5                     | 31.3                    |
| 16    | 1910.6                     | 14.8                    |
| 17    | 854.4                      | 4.0                     |
| 18    | 308.6                      | 3.6                     |
| 19    | 113.8                      | 0.0                     |

Table 1. The average number of trees consistent with 20 training examples of the $XYZ \vee AB$ concept.

## 3. Further Experimentation

For most of the problems studied, we found that on average, the smallest decision trees consistent with the training data had more error on unseen examples than slightly larger trees. We ran additional experiments to make sure that this result is not an artifact of the experimental methodology that we used, as reported in the next sections.

### 3.1 Representative Train/Test Partitions

One possible explanation for the finding of the previous section is that the smaller decision trees are formed from unrepresentative samples. For example, there are 11 positive and 21 negative examples of the concept $XYZ \vee AB$. If all or most of the examples in the training set are negative, a very small tree may be learned which would probably do very poorly on the mostly positive test set. To insure that the results are not caused by unrepresentative training sets, we eliminated all training data that was not reasonably representative. In particular, since there is a $\frac{11}{32}$ probability that a training instance is positive, a representative training set of size 20 would have about 7 positive instances. Since one standard deviation would be $\sqrt{20 * \frac{11}{32} * (1 - \frac{11}{32})}$, we eliminated from analysis those training sets with greater than 8 or fewer than 5 positive instances. Similarly, there is a 0.5 probability that each binary feature takes on a true value, so we eliminated from analysis any training data which has any feature that is true in greater than 13 or fewer than 7 instances. Figure 2 is based on the 69 of 100 trials of the $XYZ \lor AB$ concept that met this representative test. Notice that the two trials that formed the only 2 and 3 node trees were removed. Even when only the more representative training sets are considered, the average error of trees of size 4 is greater than the average error of size 5 trees.

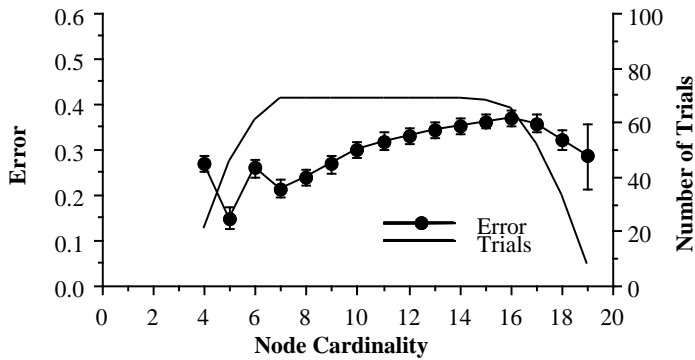

Figure 2. Error rate of consistent trees from representative training sets as a function of node cardinality.

By regrouping the results of 100 trials for the $XYZ \lor AB$ concept so that trials with the same minimum-sized trees are grouped together, a set of five curves, each associated with a subgroup, was formed (Figure 3). The intent of the grouping is to allow us to determine whether the minimum-sized trees for any given trial are on average more accurate than larger trees.

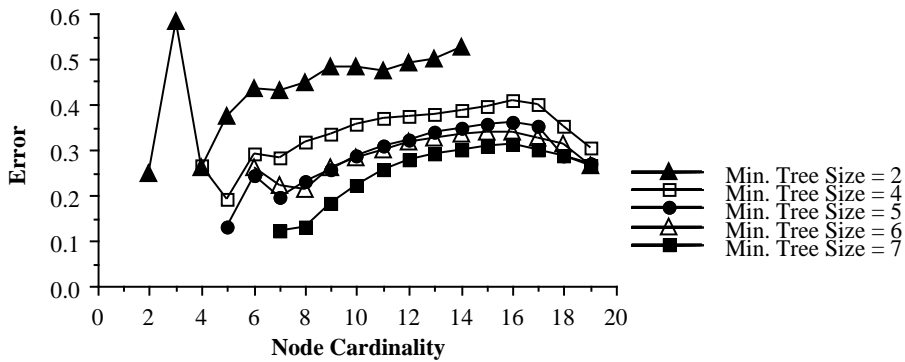

Figure 3. Error as a function of node cardinality for the $XYZ \lor AB$ concept when first grouped by minimum-sized trees built.

Note that in Figure 3, for most minimum tree sizes, error is not a monotonically increasing function of node cardinality. Furthermore, the average error of the smallest trees found is not the most accurate when the smallest tree has 4 or 6 nodes. In addition, regardless of the size of the smallest tree found, the average accuracy of trees of size 8 (the size of the smallest correct tree) rarely has the minimum average error.

Another interesting finding becomes apparent with this way of viewing the data: the average error rates of trees for training sets that allow creation of smaller consistent trees tends to be higher than for those training sets that can only form larger trees. For example, the error rate for those training sets whose minimum-sized trees have 4 nodes is higher than the error rate on trials whose minimum-sized trees has 7 nodes.

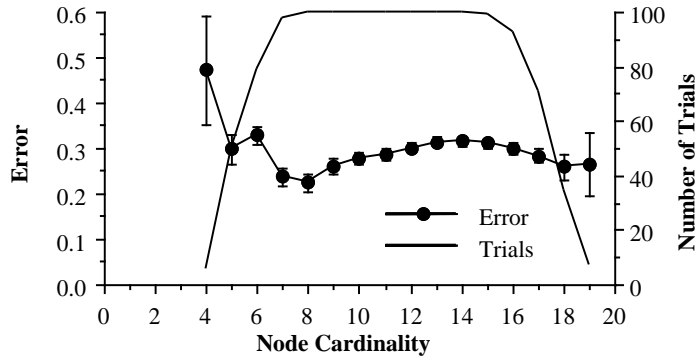

Figure 4. Error rate of consistent trees with 2 examples per leaf of some correct 8 node tree as a function of node cardinality.

The definition of representative that we used earlier in this section used global characteristics of the training data to determine representativeness. Here, we consider a more detailed view of representativeness that takes the structure of the correct concept into account. It is unreasonable to expect a decision tree learner to learn an accurate concept if there are no examples that correspond to some of the leaves of some correct decision tree. To generate training data for the next experiment, we first randomly selected one of the 72 trees with 8 nodes that is consistent with all the data. Next, for each leaf of the tree, we randomly selected two examples (if possible) to include in the training set. If a leaf only had one example, that example was included in the training set. Finally, we randomly selected from the remaining examples so that there were 20 training examples and 12 test examples. We had anticipated that with representative training sets formed in this manner, very small consistent trees would be rare and perhaps the error rate would monotonically increase with node cardinality. However, the results of 100 trials, as displayed in Figure 4, indicate the same general pattern as before. In particular, the average error of trees with 7 nodes is substantially less than the average error of those with 6 nodes. Another experiment with one randomly selected example per leaf had similar results.

## 3.2 Training Set Size and Concept Complexity

The minimum-sized decision tree for the concept $XYZ \lor AB$ has 8 tests and 9 leaves. Since the correct tree does not provide much compression[3] of a set of 20 examples used to induce the tree, one might argue that the sample used was too small for this complex a concept. Therefore, we increased the number of training examples to the maximum possible. Figure 5 plots the average error of 32 trials in which we formed all decision trees consistent with 31 examples. Each tree was evaluated on the remaining unseen example. Figure 5 shows that the smaller trees formed from samples of size 31 have more error than the slightly larger trees. Since the minimum correct decision tree has 8 nodes and the consistent trees classify all 31 training examples correctly, any decision tree with fewer than 8 nodes classifies the test example incorrectly.

---

3. The exact amount of compression provided depends upon the particular scheme chosen for encoding the training data. See (Quinlan & Rivest, 1989; Wallace & Patrick, 1993) for two such schemes.

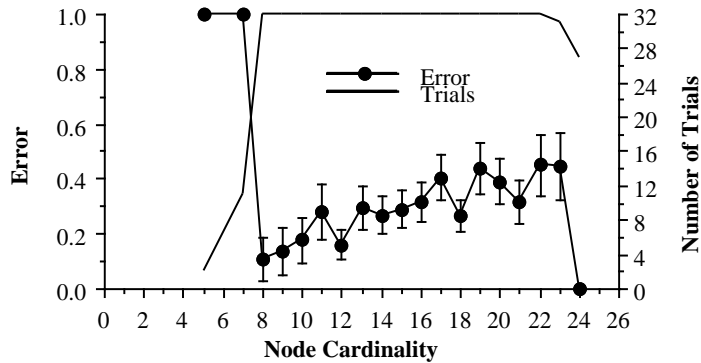

Figure 5. Error rate of consistent trees with leave-one-out testing as a function of node cardinality.

To refute further the hypothesis that the results obtained so far were based on using too small a training set for a given concept complexity, we considered two less complex concepts. In particular, we investigated a single attribute discrimination, *A* with four irrelevant features (Figure 6) and a simple conjunction, *AB* with three irrelevant features (Figure 7).

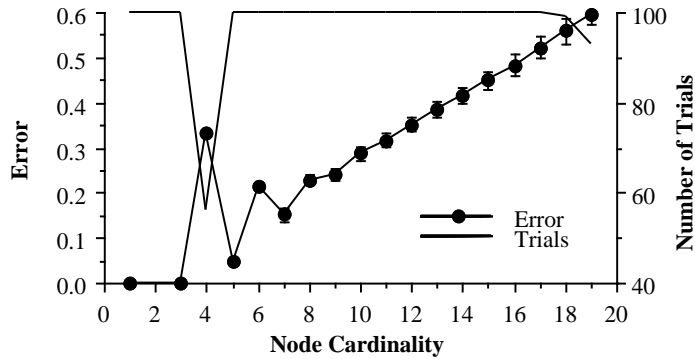

Figure 6. Error as a function of node cardinality for the single attribute discrimination *A* concept.

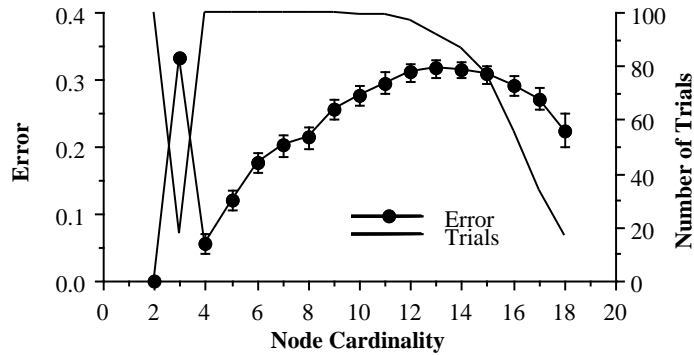

Figure 7. Error as a function of node cardinality for the simple conjunction $AB$ concept.

For each concept, 100 trials were run in which 20 examples were used for training and the remaining 12 for testing. For these simpler concepts, though the smallest trees are the most accurate, error again is not a monotonically increasing function of node cardinality.

### 3.3 Training and Testing using the Same Probability Distribution.

In our previous experiments, we used a methodology that is typical in empirical evaluations of machine learning systems: the training data and the test data are disjoint. In contrast, most theoretical work on the PAC model (Valiant, 1984) assumes that the training and test data are generated from the same probability distribution over the examples. For this section, we ran an experiment in which training and test examples were selected with replacement from the same distribution to ensure that our results were not dependent on a particular experimental methodology.

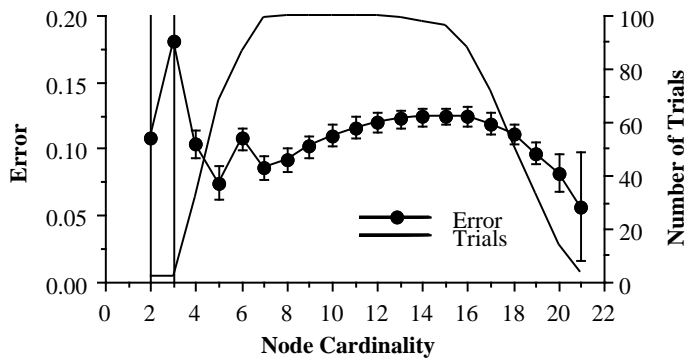

Figure 8. Error as a function of node cardinality when the training and test examples are generated by the same distribution for the $XYZ \lor AB$ concept.

Once again, the target concept was $XYZ \lor AB$. By randomly choosing 31 training examples with replacement from the set of 32 possible instances, on average approximately 20 distinct training examples are selected. Error is estimated by randomly choosing 1000 examples with replacement from the set of possible instances. Figure 8 graphs the mean error (averaged over 100 trials) as a function of node cardinality.

This testing methodology produces much smaller values for the proportion of test examples misclassified than the disjoint training and test set methodology because those test examples which also were training examples are always classified correctly. However, the same basic pattern of results is observed. Error is not at a minimum for the smallest decision trees nor at decision trees with 8 nodes (the minimum-sized correct tree). Error monotonically increases starting at trees with 7 nodes and then begins to decrease again for very large node cardinalities. Note that on some trials, it is possible to build decision trees with up to 21 nodes since some training sets contained 22 distinct examples.

## 3.4 Average Path Length

The information gain metric of ID3 is intended to minimize the number of tests required to classify an example. Figure 9 reanalyzes the data from Figure 1 by graphing average error as a function of average path length for the $XYZ \lor AB$ concept.

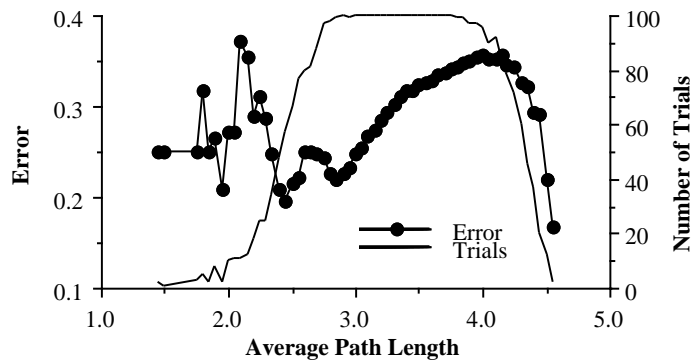

Figure 9. Error as a function of average path length for the $XYZ \lor AB$ concept.

The results are similar to those obtained when relating the number of test nodes to the error rate: error is not a monotonically increasing function of average path length. Similar analyses were performed and similar results have been obtained for other concepts which are presented in the Appendix.

## 4. The Minimum-Sized Decision Tree Policy

A designer of a learning algorithm either explicitly or implicitly must decide which hypothesis to prefer when multiple hypotheses are consistent with the training data. As Table 1 shows, there can be many consistent decision trees. Should the learner always prefer the smallest consistent decision tree? A learner that adopts this strategy can be said to be following the minimum-sized decision tree policy.

In this section, we present results from additional experiments to evaluate this policy. In particular, we gather evidence to address two related questions:

- Given any two consistent decision trees with different node cardinalities, what is the probability that the smaller decision tree is more accurate?

- Given the minimum-sized decision tree and a larger consistent decision tree, what is the probability that the smallest decision tree is more accurate?

The first question is of more interest to the current practice of decision tree induction since, for efficiency reasons, no algorithm attempts to find the smallest consistent decision tree for large data sets. Nonetheless, most algorithms are biased toward favoring trees with fewer nodes.

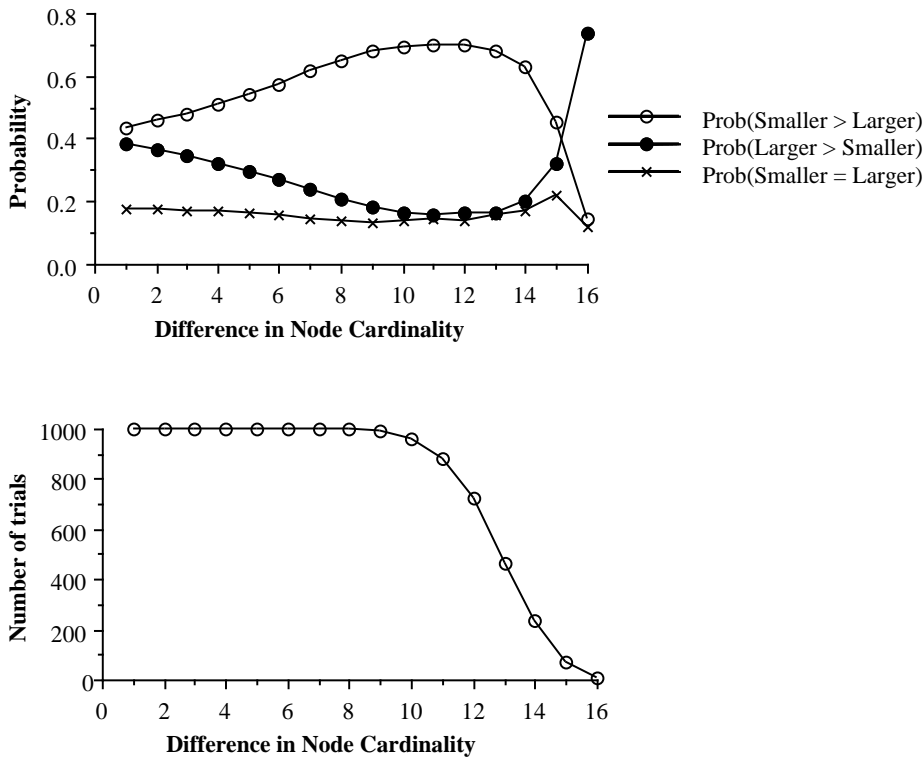

Figure 10. The probability that the accuracy of a smaller decision tree is greater than, equal to, or less than the accuracy of a larger tree as a function of the difference of node cardinalities for the $XYZ \vee AB$ concept (upper). The number of trials out of 1000 on which at least 2 trees had a given difference in node cardinality (lower).

To address the question of whether a learner should prefer the smaller of two randomly selected consistent trees, we ran 1000 trials of learning the concept $XYZ \vee AB$ from 20 training examples. For each trial, we recorded the node cardinality and accuracy (on the 12 test examples) of every consistent tree. For each pair of consistent trees (with different node cardinalities), we computed the difference in node cardinality and indicated whether
the accuracy of the smaller tree was greater than, equal to, or less than the accuracy of the
larger tree. From this data, we computed the observed probability that one decision tree
was more accurate than another as a function of the difference in node cardinalities (see
Figure 10 upper). The graph shows that on this concept, the probability that the smaller
of two randomly chosen consistent decision trees will be more accurate is greater than the
probability that the larger tree will be more accurate. Furthermore, the probability that
the smaller tree is more accurate increases as the difference in node cardinality increases.
An exception to this trend occurs for very large differences in node cardinality. However,
as Figure 10 lower shows, these exceptions are quite rare. Consistent decision trees whose
node cardinalities differed by 16 were found in only 6 of the 1000 trials.[4] The results of
this experiment indicate that on average, a learner that prefers the smaller of two randomly
selected decision trees has a higher probability of being more accurate on this concept than
a learner that selects the larger tree.

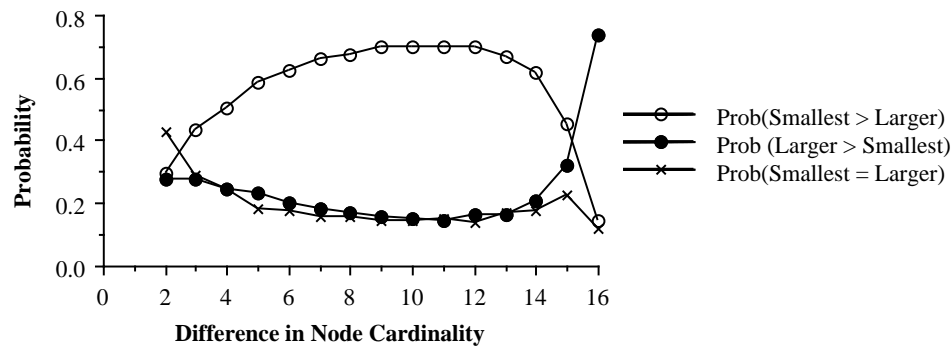

Figure 11. The probability that the accuracy of a minimum-sized decision is greater than,
equal to, or less than the accuracy of a larger tree as a function of the difference of node
cardinalities for the $XYZ \lor AB$ concept.

To address the question of whether a learner should prefer the smallest consistent deci-
sion over a randomly selected consistent tree with more test nodes, we reanalyzed the data
from the previous experiment. Figure 11 graphs the observed probability that a consistent
decision tree with the minimum node cardinality is more accurate than a larger tree as a
function of the difference in node cardinalities between the two trees. The graph shows that
a learner that chooses randomly among the consistent decision trees with minimum node
cardinalities is more likely to find a tree that is more accurate than a learner that randomly
selects among larger trees.[5]

Figure 11 clearly shows that for this particular concept, preferring the minimum-sized
decision tree policy is on average a better policy than preferring a decision tree that is any

---

fixed size larger than the smallest decision tree. However, it is not clear that the minimum-sized decision tree is the best possible policy for this concept. Indeed, by looking at the data from Figure 3, it is apparent that a better strategy for this concept would be to find the minimum-sized tree and then decide whether to return the minimum-sized tree or a tree of a different node cardinality as a function of the node cardinality of the minimum-sized consistent tree. Table 2 shows which node cardinality has the highest probability of being most accurate as a function of the minimally sized tree, together with the number of trials (out of 1000) on which the minimum-sized tree had a particular node cardinality.

| Minimum Node Cardinality | Preferred Node Cardinality | Number of Trials |
|:---:|:---:|:---:|
| 2 | 2 | 49 |
| 3 | 5 | 17 |
| 4 | 5 | 300 |
| 5 | 5 | 351 |
| 6 | 8 | 211 |
| 7 | 8 | 71 |
| 8 | 8 | 1 |

Table 2. A policy of returning a larger decision tree as a function of the minimum-sized tree for the $XYZ \lor AB$ concept.

Figure 11 provides some of the data that illustrates that the policy in Table 2 will perform better than preferring the minimum-sized decision tree on this concept. Figure 12 graphs the observed probability that a consistent decision tree with a minimum node cardinality of 5 (upper), 6 (middle), or 7 (lower) is more accurate than a larger tree as a function of the difference in node cardinalities between the two trees. The graph shows that when the minimum-sized decision tree has 5 nodes, the probability that a larger tree is more accurate is less than the probability that the smaller tree is more accurate for all node cardinalities. This is particularly interesting because it shows that giving a decision tree learner the size of the correct tree and having the decision tree learner produce an hypothesis of this size is not the best strategy for this concept. However, when the smallest consistent tree has 6 nodes, there is a 0.560 probability that a randomly chosen tree with 8 nodes will be more accurate and a 0.208 probability that a tree with 8 test nodes will have the same accuracy. In addition, when the minimum-sized tree has 7 test nodes, the probability that a tree with 8 nodes is more accurate is 0.345 while the probability that it is less accurate is 0.312.

Note that we do not believe that the policy in Table 2 is uniformly superior to preferring the minimum-sized decision tree. Rather, there is probably some interaction between the complexity of the concept to be learned, the number of training examples, and the size of the smallest consistent decision tree. Furthermore, a learner should not be tuned to learn a particular concept, but should perform well on a variety of concepts. Clearly, if extremely simple concepts are to be learned sufficiently frequently, the minimum-sized decision tree policy will be better than the policy in Table 2. Indeed, the minimum-sized decision tree policy would work well on the simple concepts $A$ and $AB$ discussed in Section 3.2. However, if simple concepts are rarely encountered, there may be better policies. The best policy must depend upon the distribution of concepts that are encountered. Clearly, if the only concept to be learned is $XYZ \lor AB$, the best policy would be to ignore the training data and return the decision tree representation for $XYZ \lor AB$. It may be that Occam's razor should be viewed as a philosophical statement about the distribution of concepts one is likely to encounter. Occam's razor has not been shown to be a guarantee that when learning a complex concept, the simplest hypothesis consistent with the data is likely to be more accurate than the randomly-chosen more complex hypothesis consistent with the training data.

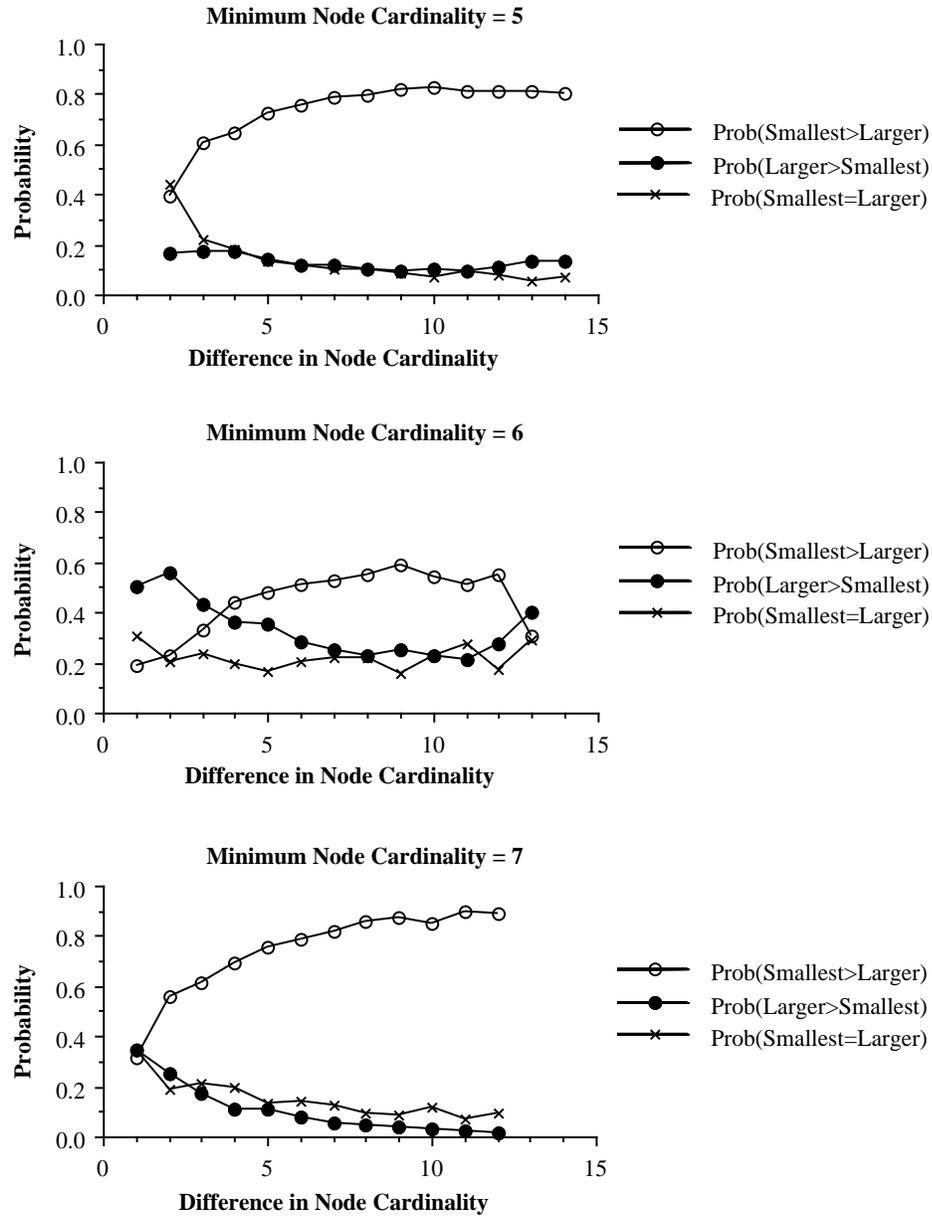

Figure 12. The probability that the accuracy of a minimum-sized decision tree is greater than, equal to, or less than the accuracy of a larger tree as a function of the difference of node cardinalities for the $XYZ \lor AB$ concept when the minimum-sized decision tree has 5 (upper), 6 (middle), or 7 (lower) test nodes.

## 5. Analysis

Schaffer (1992, 1993) presents a series of experiments on overfitting avoidance algorithms. Overfitting avoidance algorithms prefer simpler decision trees over more complex ones, even though the simpler decision trees are less accurate on the training data, in hopes that the trees will be more accurate on the test data. Schaffer shows that these overfitting avoidance algorithms are a form of bias. Rather than uniformly improving performance, the overfitting avoidance algorithms improve performance on some distributions of concepts and worsen performance on other distributions of concepts.

The results of our experiments go a step further than Schaffer's. We have shown that for some concepts, the preference for simpler decision trees does not result in an increase in predictive accuracy on unseen test data, even when the simple trees are consistent with the training data. Like Schaffer, we do not dispute the theoretical results on Occam's razor (Blumer, Ehrenfeucht, Haussler, & Warmuth, 1987), minimum description length (Quinlan & Rivest, 1989; Muggleton et al., 1992), or minimizing the number of leaves of a decision tree (Fayyad & Irani, 1990). Rather, we point out that for a variety of reasons, the assumptions behind these theoretical results mean that the results do not apply to the experiments reported here. For example, (Blumer et al., 1987) indicates that if one finds an hypothesis in a sufficiently small hypothesis space (and simpler hypotheses are one example of a small hypothesis space) and this hypothesis is consistent with a sufficiently large sample of training data, one can be fairly confident that it will be fairly accurate on unseen data drawn from the same distribution of examples. However, it does not say that on average this hypothesis will be more accurate than other consistent hypotheses not in this small hypothesis space.

The (Fayyad & Irani, 1990) paper explicitly states that the results on minimizing the number of leaves of decision trees are worst case results and should not be used to make absolute statements concerning improvements in performances. Nonetheless, informal arguments in the paper state: "This may then serve as a basis for provably establishing that one method for inducing decision trees is better than another by proving that one algorithm always produces a tree with a smaller number of leaves, given the same training data." Furthermore, other informal arguments imply that this result is probabilistic because of the existence of "pathological training sets." However, as we have shown in Figures 2 and 4 (as well as a reanalysis of the mux6 data in the Appendix), eliminating pathological (i.e., unrepresentative) training sets does not change the qualitative result that on some concepts, the smaller trees are less accurate predictors than slightly larger trees.

## 6. Conclusion

We have reported on a series of experiments in which we generated all decision trees on a variety of artificial concepts and two naturally occurring data sets. We found that for many of the concepts, the consistent decision trees that had a smaller number of nodes were less accurate on unseen data than the slightly larger ones. These results do not contradict existing theoretical results. Rather, they serve to remind us to be cautious when informally using the intuitions derived from theoretical results on problems that are not covered by the theorems or when using intuitions derived from worst-case results to predict average-case performance.

We stress that our results are purely experimental. Like the reader, we too would be pleased if there were theoretical results that indicated, for a given sample of training data, which decision tree is likely to be most accurate. However, it is not clear whether this can be done without knowledge of the distribution of concepts one is likely to encounter (Schaffer, 1994).

We also note that our results may be due to the small size of the training sets relative to the size of the correct tree. We tried to rule out this possibility by using larger training sets (31 of the 32 possible examples) and by testing simpler concepts. For the simpler concepts, the smallest decision trees were the most accurate, but error did not monotonically increase with node cardinality. Since most decision tree learners that greedily build decision trees do not return the smallest decision tree, our results may be of practical interest even for simple concepts. In the future, experiments with more features and more examples could help to answer this question, but considerably more complex problems cannot be handled even by future generations of parallel supercomputers. In addition, we note that in our experiments, we did not build decision trees in which a test did not partition the training data. This explains why we found relatively few extremely large decision trees and may explain why very large trees made few errors. To our knowledge, all decision tree algorithms have this constraint. However, the theoretical work on learning does not make use of this information. We could rerun all of our experiments without this constraint, but we would prefer that some future theoretical work take this constraint into account.

Although we have found situations in which the smallest consistent decision tree is not on average the most accurate and cases in which there is a greater than 0.5 probability that a larger decision tree is more accurate than the smallest, we believe that learning algorithms (and people) with no relevant knowledge of the concept and no information about the distribution of concepts that are likely to be encountered should prefer simpler hypotheses. This bias is appropriate for learning simple concepts. For more complex concepts, the opposite bias, preferring the more complex hypotheses, is unlikely to produce an accurate hypothesis (Blumer et al., 1987) and (Fayyad & Irani, 1990) due to the large number of consistent complex hypotheses. We believe that the only way to learn complex hypotheses reliably is to have some bias (e.g., prior domain knowledge) which favors particular complex hypotheses such as combinations of existing hypotheses learned inductively as in OCCAM (Pazzani, 1990). Indeed, (Valiant, 1984) advocates a similar position: "If the class of learnable concepts is as severely limited as suggested by our results, then it would follow that the only way of teaching more complex concepts is to build them up from simpler ones."

## Acknowledgements

We thank Ross Quinlan, Geoffrey Hinton, Michael Cameron-Jones, Cullen Schaffer, Dennis Kibler, Steve Hampson, Jason Catlett, Haym Hirsh, Anselm Blumer, Steve Minton, Michael Kearns, Tom Dietterich, Pat Langley, and David Schulenburg for commenting on various aspects of this research. The research reported here was supported in part by NSF infrastructure grant number MIP-9205737, NSF Grant INT-9201842, AFOSR grant F49620-92-J-0430, and AFOSR AASERT grant F49620-93-1-0569.

## Appendix A. Experiments on Additional Problems

In this appendix, we provide data on experiments which we ran on additional problems. The experiments show that the basic findings in this paper are not unique to the artificial concept, $XYZ \lor AB$.

### Mux6

The multiplexor concept we consider, mux6, has a total of 8 binary features. Six features represent the functionality of a multiplexor and 2 features are irrelevant. The minimum sized tree has 7 nodes. This particular concept was chosen because it is difficult for a top-down inductive decision tree learner with limited look ahead to find a small hypothesis (Quinlan, 1993). On each trial, we selected 20 examples randomly and tested on the remaining examples. Since most of the computational cost of building consistent trees is for larger node cardinalities and we are primarily interested in trees with small node cardinalities, we only computed consistent trees with up to 10 nodes for 10 trials and up to 8 nodes for 340 trials. Figure 13 presents the average error as a function of the node cardinality for these trials. This graph again shows that average error does not monotonically increase with node cardinality. Trees of 4 nodes are on the average 4% less accurate than trees of 5 nodes.

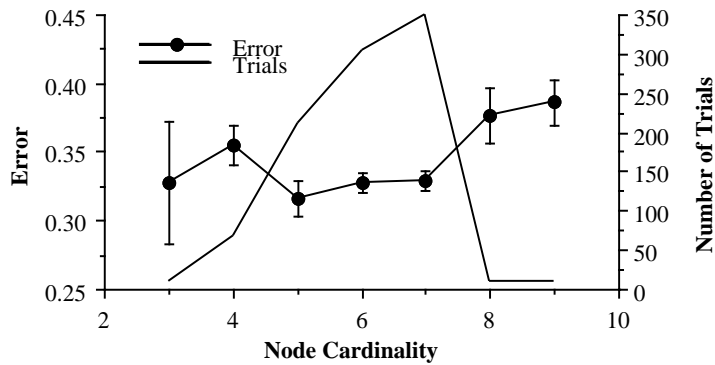

Figure 13. Error as a function of node cardinality for the mux6 concept.

## Lenses

The lenses domain has one 3-valued and three binary features, three classes, and 24 instances. Since the lenses domain has one non-binary feature, trees with a range of leaf cardinalities are possible for a particular node cardinality. The minimum-sized tree has 6 nodes and 9 leaves. Separate analyses for leaf and node cardinalities were performed. We used training set sizes of 8, 12, and 18 for this domain, built all consistent trees, and measured the error rate on all unseen examples.

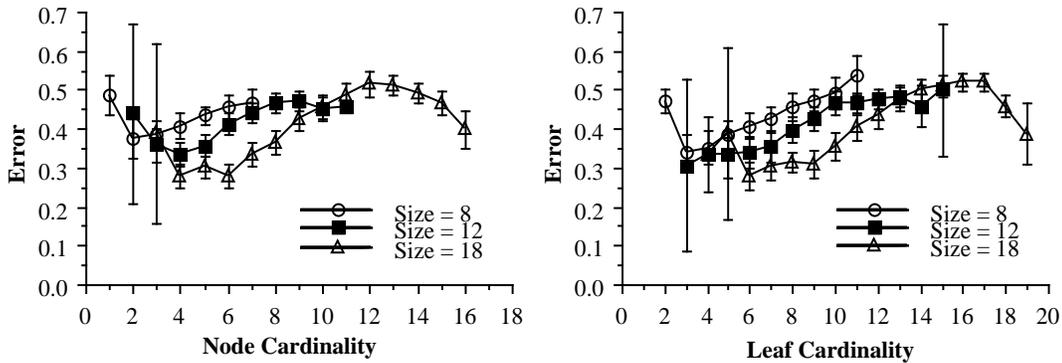

Figure 14. Error as a function of node cardinality (left) and error as a function of leaf cardinality (right).

Figure 14 (left) shows the error as a function of the node cardinality for the 3 training set sizes averaged over 50 trials. These curves indicate that the smallest consistent trees are not always the most accurate. When observing the larger node cardinalities for the training set sizes 12 and 18, error monotonically decreases with increasing node cardinality. Similar statements can be said for the curve in Figure 14 (right), which relates average error as a function of leaf cardinality.

## Shuttle Landing

The shuttle landing domain has four binary and two 4-valued features, two classes, and 277 instances. The minimum-sized consistent tree has 7 nodes and 14 leaves. We used training sets of size 20, 50, and 100 for the shuttle domain, generating all consistent decision trees with fewer than 8, 10, and 12 nodes, and measured the error of these trees on all unseen examples. Figure 15 presents the error as a function of leaf cardinality, averaged over 10 trials. For this domain, there is a monotonically increasing relationship between node cardinality and error.

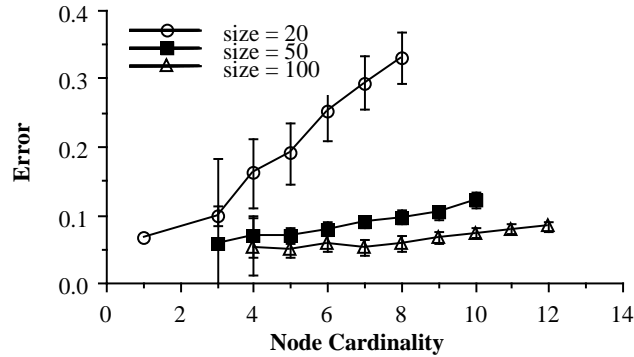

Figure 15. Error as a function of node cardinality for the Shuttle concept.